\documentclass[10pt,twocolumn,letterpaper]{article}

\usepackage[pagenumbers]{cvpr} 
\usepackage{float}
\usepackage[utf8]{inputenc}

\usepackage{graphicx}
\usepackage{amsmath}
\usepackage{booktabs}
\usepackage{multirow}
\usepackage{algorithm}
\usepackage{algpseudocode}
\usepackage{xcolor}
\usepackage{amssymb}
\usepackage{longtable}
\usepackage{bbding} 
\usepackage[pagebackref,breaklinks,colorlinks]{hyperref}

\usepackage[capitalize]{cleveref}
\crefname{section}{Sec.}{Secs.}
\Crefname{section}{Section}{Sections}
\Crefname{table}{Table}{Tables}
\crefname{table}{Tab.}{Tabs.}

\def\cvprPaperID{16894} 
\def\confName{CVPR}
\def\confYear{2024}

\begin{document}

\title{Fuse after Align: Improving Face-Voice Association Learning via Multimodal Encoder}

\author{Chong Peng\\
Carnegie Mellon University\\
{\tt\small chongp@andrew.cmu.edu}
\and
Liqiang He\\
Tencent AI Lab\\
{\tt\small andylqhe@tencent.com}
\and
Dan Su\\
Tencent AI Lab\\
{\tt\small dansu@tencent.com}
}
\maketitle

\begin{abstract}
    Today, there have been many achievements in learning the association between voice and face. However, most previous work models rely on cosine similarity or L2 distance to evaluate the likeness of voices and faces following contrastive learning, subsequently applied to retrieval and matching tasks. This method only considers the embeddings as high-dimensional vectors, utilizing a minimal scope of available information. This paper introduces a novel framework within an unsupervised setting for learning voice-face associations. By employing a multimodal encoder after contrastive learning and addressing the problem through binary classification, we can learn the implicit information within the embeddings in a more effective and varied manner. Furthermore, by introducing an effective pair selection method, we enhance the learning outcomes of both contrastive learning and the matching task. Empirical evidence demonstrates that our framework achieves state-of-the-art results in voice-face matching, verification, and retrieval tasks, improving verification by approximately 3\%, matching by about 2.5\%, and retrieval by around 1.3\%.
\end{abstract}

\section{Introduction}
\label{sec:intro}
Studies in neuroimaging have indicated a connection between voice and face, not only at a perceptual level but also in the process of person recognition\cite{Smith16,Kamachi03}. This suggests that we can instinctively visualize their appearance upon hearing someone's voice, even without seeing them. Conversely, we can also construct an idea of a stranger's voice based on their facial features. In recent years, researchers have attempted to explore this relationship further using machine learning and deep learning methods\cite{kim2018learning,Pins,SVHF,distangle,nagrani2018seeing,wen2019face,cmpc,fop,selflifting}, enabling computational models to recognize faces and voices and associate them with the same identity. This area of research is not only fascinating from a scientific perspective but also has potential applications in various fields, such as virtual reality, criminal investigations, and multimodal information retrieval. By formulating this as a voice-face association problem, we can investigate the cross-modal matching, verification, and retrieval aspects to better understand the complex relationship between voice and face.

In recent years, numerous outstanding works have emerged in this field. SVHF\cite{SVHF} treats the Matching task as a classification task and achieves relatively good results in both face-to-voice and voice-to-face Matching. With the rapid development of deep learning, some works have further demonstrated breakthroughs in solving such problems using metric learning methods. In subsequent research, many studies have attempted to enhance the model's performance and robustness through methods such as noise learning\cite{wen2021seeking} and disentangled representation learning\cite{distangle}. At the same time, studies have also tried to improve the loss function to enhance the effectiveness of contrastive learning, with the expectation of achieving better modality alignment results\cite{cmpc,fop}.

Although previous works have achieved relatively good results, there are still three major issues. 

\begin{itemize}
    \item [\textbullet]They all rely on cosine similarity or L2 distance for tasks such as Verification and Matching. However, not all the correlation information contained in the embedding can be measured by these two methods.

    \item [\textbullet] Their approach only involves a single training objective, Contrastive Learning. In addition to the potential insufficiency in model training due to the limitations of Contrastive loss\cite{contrastiveloss} or triplet loss\cite{triplet}, this approach may also prevent the model from learning more diverse and rich relationships between samples.

    \item [\textbullet]The selection of training samples used are not diverse enough and have low learning difficulty, resulting in insufficient generalization performance and lack of robustness in the model.
\end{itemize}

\begin{figure}[t]
  \centering
   \includegraphics[width=1\linewidth]{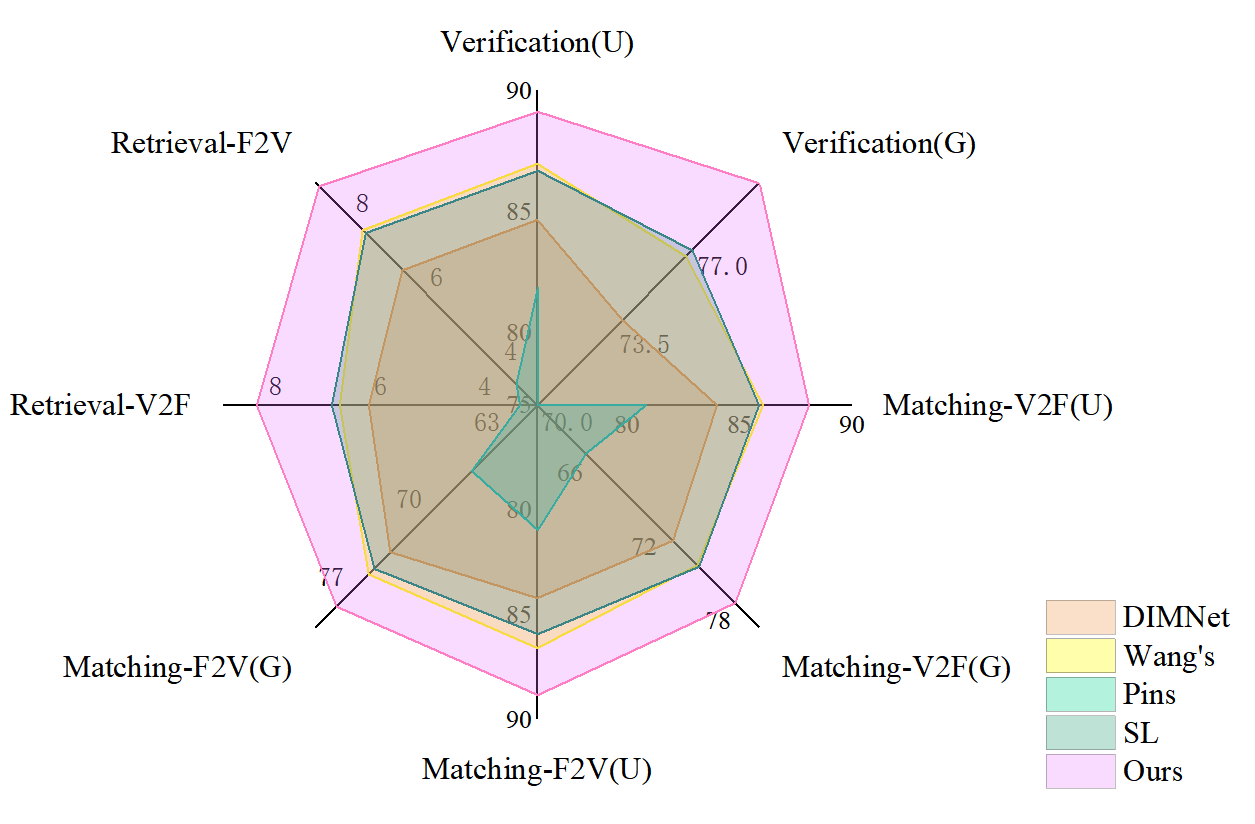}
   \caption{FAA achieves state-of-the-art performance on various tasks when comparing to previous works}
   \label{fig:onecol}
\end{figure}

\begin{figure*}[t]
  \centering
   \includegraphics[width=1\linewidth]{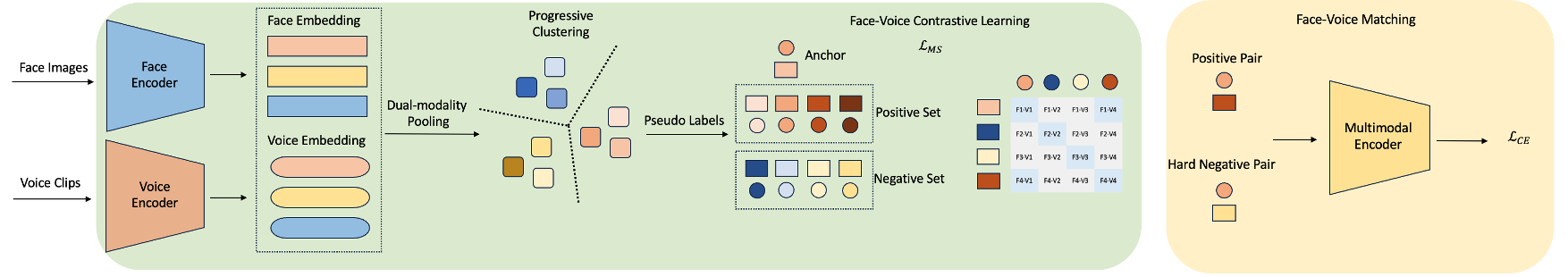}
   \caption{An overview of the proposed method: Face and voice data without identity information are input into a face and voice encoder, respectively. The resulting embeddings are then subjected to dual-modality pooling and clustering to generate pseudo-labels. We introduce a face-voice contrastive learning for modality alignment and propose a face-voice matching to train the Multimodal encoder. The aim is to use it to learn the relationships between modalities. The entire training process iteratively repeats the clustering and metric learning steps.}
   \label{fig:onecol}
\end{figure*}

To address the first issue, we propose using a multimodal encoder to perform cross-modal learning. It can fuse face and voice features while utilizing self-attention to enhance the cross-modal learning capability. We train this multimodal encoder using the Face-Voice Matching loss, allowing it to focus on learning the matching relationships between samples. In addition, the classification layer corresponding to this task can also be directly applied to downstream tasks such as Verification.

To address the second issue, we combine modality alignment with task learning. Following\cite{selflifting}, We employ the multi-similarity loss function\cite{wang2020multisimilarity} to implement Contrastive Learning, which can consider a more significant number of samples compared to the Contrastive loss function and other loss functions, allowing for a more comprehensive understanding of the relationships between samples. Concurrently, we incorporate Face-Voice Matching to enable the multimodal encoder to learn cross-modal relations.

To tackle the third issue, we attempt to use an effective method for pair selection, combining diverse positive pair selection\cite{tao2022selfsupervised} with hard negative mining\cite{albef}. We apply progressive clustering to find diverse positive pairs. When performing the Face-Voice Matching task, we apply the Global Hard Negative Mining method to search for hard negative pairs. These two methods can improve the quality of the found positive pairs and negative pairs, making Contrastive Learning and Face-Voice Matching more challenging and resulting in a more robust model with better generalization performance.

In summary, we propose the Fuse after Align (FAA) framework. Our main contributions can be summarized as follows:

\begin{itemize}
    \item [\textbullet]We propose using a multimodal encoder to learn cross-modal relations from a deeper and more diverse perspective.

    \item [\textbullet]We propose employing mixed training objects, achieving both better modality alignment results and direct cross-modality learning.

    \item [\textbullet]We propose an effective pair selection method to enhance the diversity of training samples and thereby improve the model's ability to handle hard samples.
\end{itemize}

\section{Related Work}

\subsection{Learning face-voice association}
In recent years, the exploration of voice-face association learning has garnered significant interest among researchers. The seminal work by SVHF\cite{SVHF} introduced the face-voice matching task, conceptualizing it as a binary classification problem that employs cross-entropy loss to model the association between faces and voices. Subsequently, researchers have tended to address this problem through cross-modal learning. Efforts in this domain aim to derive embeddings for voices and faces and then manipulate these embeddings to minimize distances for the same identities and maximize distances for different identities.

Various studies have implemented distinct loss functions to refine the learning process. For instance, the use of contrastive loss in Pins\cite{Pins}, triplet loss in \cite{kim2018learning}, N-pair loss in \cite{Horiguchi,wen2021seeking}, and multi-similarity loss in \cite{selflifting} have all contributed to substantial advancements in modality alignment. Furthermore, FOP \cite{fop} introduced an orthogonality constraint loss to enhance intra-identity closeness while ensuring clear separability between different identities. SSNet \cite{ssnet} employed center loss to define centers for all classes, including faces and voices, and worked on minimizing the distances between each center.

In addition to refining loss functions, substantial work has been done to enhance the quality of training data, develop better-structured networks, and devise comprehensive learning strategies. Nawaz et al. \cite{nawaz2019deep} proposed an end-to-end network capable of mapping multiple modalities into a shared latent space without relying on pairwise or triplet data. CMPC\cite{cmpc} utilizes contrastive methods to mitigate the impact of false negatives and deviations in positives. Wen et al. \cite{wen2021seeking} apply sample re-weighting to filter out idiosyncratic identities and use both implicit and explicit modality alignments to deepen the understanding of the relationship. Tao et al. \cite{tao2022selfsupervised} explore progressive clustering to discover a variety of positive pairs, and Chen et al. \cite{guangyulocalglobal} advocate for an optimization objective that combines local and global contrastive elements to enhance the learning of voice-face representations. Approaches such as disentangled representation learning\cite{distangle} and adversarial learning\cite{adver1,adver2} have also been employed to study the voice-face relationship.

Despite these innovations, much of the existing research has centered on optimizing the similarity metrics—such as cosine similarity or L2 distance—between face and voice embeddings for identical identities and the opposite for different ones, particularly in evaluating downstream tasks. FOP \cite{fop} attempts to fuse face and voice embeddings to improve alignment, yet the evaluation still hinges on measuring the L2 distance between the two modalities. In contrast, our approach leverages neural networks to enhance understanding of these correlations.

\section{Methods}
We aim to obtain a feature extraction model that can effectively map facial and voice features into the same vector space, bringing features of the same identity closer together while pushing features of different identities further apart. Doing so enables us to perform tasks such as face-voice/voice-face matching, verification, and retrieval with great accuracy. To achieve this goal, we employ the FAA framework.

Section 3.1 introduces a progressive clustering method, which plays a crucial role in helping us achieve high-quality self-supervised learning by assigning pseudo-labels and obtaining diverse, high-quality clustering results. Moving on to Section 3.2, we discuss the training objectives in detail. Notably, the modality alignment loss effectively reduces the semantic gap, as described in Section 3.2.1.

Furthermore, Section 3.2.2 elaborates on how the multimodal encoder is trained by the matching loss, resulting in a comprehensive understanding of the relationship between facial and voice features.

\begin{algorithm}
\caption{Training with FAA}
\begin{algorithmic}[1]
\renewcommand{\algorithmicrequire}{\textbf{Input:}}
\renewcommand{\algorithmicensure}{\textbf{Output:}}

\Require Unlabeled video dataset $D$; initial cluster number $C$; initial model $\phi(\cdot, \theta)$; training iteration $T$; batch size $M$; epoch $P$.
\Ensure Model parameter $\theta$.
\State \textcolor[RGB]{238,178,121}{\textbf{Training Data Preparation}}

\State Extract face images and voice clips $\{x_f^n, x_v^n\}^{N}_{n=1}$  from video dataset $D$;
\For{$p=1$ to  $P$}

    \State {\textcolor[RGB]{238,178,121}{\textbf{Clustering Step}}}
    \State Extract video feature $\{e^n\}^N_{n=1}$ with Eq. (1);
    \State Clustering $\{e^n\}^N_{n=1}$ into $C$ clusters;
    \State Assign pseudo-label $\{y^n\}_{n=1}^N$ to videos;
    \State {\textcolor[RGB]{238,178,121}{\textbf{Metric Learning Step}}}
    \State Init labeled voice-face set $D'=\{x_f^n, x_v^n, y^n\}^{N}_{n=1}$;
    \For{mini-batch$(D',M)$ in $D'$}
        \State Extract feature $\{e_f^m, e_v^m\}^{M}_{m=1}$;
        \State Calculate $\mathcal{L}_{MS}$ with Eq. (4);
        \State Calculate $\mathcal{L}_{CE}$ with Eq. (5);
        \State Update model parameters $\theta$ with Eq. (6).
    \EndFor
    \State Calculate performance on valid set $res=val(\theta)$.
    \If{$res$ can't improve}
        \State $C \gets   C/2$.
    \EndIf
\EndFor
\end{algorithmic}
\end{algorithm}

\subsection{Clustering}
\subsubsection{Dual-modality Pooling}
Our approach, which clusters videos by extracting voice clips and faces, offers significant benefits. By utilizing both voice and face modalities, we concatenate the pooled voice features and pooled face features to represent each video. This approach not only enhances the accuracy and diversity of clustering results, but also improves efficiency by reducing the number of samples required for clustering, compared to traditional methods that use single images or voice clips.

Specifically, the voice clips $v^i$ and face images $f^i$ are extracted from one video $z_i$, and the features of face images and voice clips are obtained through face encoder $\theta_f$ and voice encoder $\theta_v$. Then the video representation $Emb_i$ is calculated through intra-modality
feature pooling and inter-modality concatenation, as shown in Formula 1:

\begin{equation}
  Emb_i=\frac{1}{n}\sum_{j=1}^{n}\theta_f(f_j^i)\oplus \frac{1}{m}\sum_{j=1}^{m}\theta_v(v_j^i),
  \label{eq:important}
\end{equation}where $\oplus$ stands for vector concatenation, $n$ and $m$ denotes the number of face images and voice clips extracted from video $z_i$.

\subsubsection{Progressive clustering\cite{tao2022selfsupervised}}
We adopt a simple yet effective algorithm, k-means, to implement clustering. Initially, we use a high number of clusters $C$ and progressively reduce the number of clusters. In a dataset without speaker labels, we set $C$ as the number of training video clips. During training, as soon as the performance for the validation set stops improving, we halve the number of clusters $C$ and repeat the clustering process. As C decreases, we force more video clips into one cluster, thus making more diverse positive counterparts available.

\subsection{Training Obejectives}
In this section, we propose two training objectives to achieve modality alignment while obtaining a powerful multimodal encoder. Compared to directly using cosine similarity or L2 distance for a sample comparison, this multimodal encoder can better learn the relationship between the two modalities, ultimately leading to improved performance in verification or matching tasks.
\subsubsection{Face-Voice Contrastive Learning}
Face-Voice Contrastive Learning aims to learn better unimodal representations before fusion. We employ the multi-similarity loss to achieve modality alignment, which is more effective at learning the relationships between samples than the contrastive loss, which considers constraints between two samples, or the Triplet loss, which considers three samples.

The pseudo-labels generated $y$ through the clustering step are used as the supervisory information. The multi-similarity loss is calculated as follows: 
\begin{itemize}
  \item [1)]

    We extract $N$ face-voice tuples from $N$ videos and obtain embeddings through the face and voice encoder. We use $B=\{(x_i,y_i)\}_{i=1}^{2N}$ to represent the embeddings in this batch, where $x_i$ denotes either a face or voice embedding.
  \item [2)]
  Build the positive set: $P_i=\{x_j|y_j=y_i,j \neq i\}$ and the negative set:$N_i=\{x_j|y_j\neq y_i\}$.

  \item [3)]
  Assuming $x_i$ is an anchor, we next mine informative negative pairs $N_i$ where a negative pair$\{x_i,x_j\}$ is selected if their cosine similarity $S_{ij}$ satisfies the condition:
  
    \begin{equation}
      S_{ij}^- > \min_{y_m=y_i}S_{im}-\epsilon.
      \label{eq:important}
    \end{equation}
if $\{x_i,x_j\}$ is a positive pair, $S_{ij}$ has to satisfy the condition:
    \begin{equation}
      S_{ij}^+ < \min_{y_m\neq y_i}S_{im}+\epsilon,
      \label{eq:important}
    \end{equation}
    where $\epsilon$ is a given margin, $S_{im}$ is the cosine similarity between $x_i$ and $x_m$.
  \item[4)]
  Then we calculate multi-similarity (MS) loss as follows:
  \begin{equation}
  \begin{array}
  {r}\mathcal{L}_{MS}=\frac{1}{2N} \sum\limits_{i=1}^{2N}\{\frac{1}{\alpha} \log [1+\sum\limits_{k \in {P}_i} e^{-\alpha(S_{i m}-\lambda)}] \\ +\frac{1}{\beta} \log [1+\sum\limits_{k \in {N}_i} e^{\beta(S_{i m}-\lambda)}]\} ,
  \end{array}
  \label{eq:important}
  \end{equation}
where $\alpha$, $\beta$, $\lambda$ are fixed hyperparameters.

\end{itemize}

\subsubsection{Face-Voice Matching}
Face-Voice Matching predicts whether a pair of face images and voice clips is from the same identity (positive) or not (negative). We use the pooled multimodal encoder's output embedding as the joint representation of the face-voice pair, adding a fully connected layer and a softmax layer, resulting in a binary classification probability. If the given pair is from the same identity, it should be classified as 1; if the given pair is not from the same identity, it should be classified as 0. Assuming that $\hat{y}$ represents the probability of the sample pair being a positive case, the calculation of the loss function is as follows:
\begin{equation}
\mathcal{L}_{CE}=-\frac{1}{N}\sum_{i=1}^N(y_i\cdot \log(\hat{y}_i)+(1-y_i)\cdot \log (1-\hat{y}_i))
\label{eq:important}
\end{equation}

We employ the Global Hard Negative Mining method to select negative samples, which involves collecting all training samples across GPUs and mining hard sample pairs from the collected data.

The full training objective of FAA is :
\begin{equation}
\mathcal{L}=\delta\cdot\mathcal{L}_{MS}+(1-\delta)\cdot\mathcal{L}_{CE},
\label{eq:important}
\end{equation}
where $\delta\in(0,1)$  is a hyper parameter.

\section{Experiments}
\subsection{Dataset}
We train our model on the VoxCeleb dataset\cite{voxceleb}, which has been widely used in previous works to evaluate the performance of frameworks. Additionally, we use face images extracted from VoxCeleb videos provided by Pins, combined with the audio clips from VoxCeleb, to create our face-voice dataset. This dataset comprises 153K audio clips and 1.2M face images of 1225 identities. It is worth noting that we do not use speaker identity labels during training. We partition the dataset into train/validation/test sets, and the dataset's statistics are shown in Table 1.

\begin{table}
  \centering
  \begin{tabular}{c|cccc}
    \toprule
     & train & validation & test & total \\
    \midrule
    video & 16,650 & 2045& 3428& 22123\\
    audio & 113,322&  14,182& 21,850& 149,354 \\
    face & 905,530&  108,951& 170,330& 1,184.811\\
    id & 924 & 112& 189& 1225\\
    \bottomrule
  \end{tabular}
  \caption{Dataset partition}
  \label{tab:example}
\end{table}

\begin{table*}[h]
  \centering
  \caption{Results(\%) on Verification, 1:2 Matching and Retrieval. V2F: from voice to face; F2V: from face to voice; U: gender unrestricted; G: same-gender restricted; Veri is short for Verification. The best results of all models and competitors are indicated in bold and underlined.
}
  \label{tab:example}
  \begin{tabular}{c|c|cc|c|cccc|cc}
\toprule
  \multirow{2}{*}{Methods}&\multirow{2}{*}{Kind of Sup.} & \multicolumn{2}{c}{Veri.(AUC)}&Veri.(EER)& \multicolumn{4}{c}{ 1:2 Matching(ACC) } & \multicolumn{2}{c}{ Retrieval(mAP) } \\
  && U & G &U& V2F(U) & V2F(G) & F2V(U) & F2V(G) & V2F & F2V \\
\midrule
  DIMNet\cite{dimnet} &Supervised & $84.22 $ & $74.01 $ &$23.0$& $84.02 $ & $72.97 $ & $84.13 $ & $73.16 $ & $6.21 $ & $6.35 $ \\
  SSNet\cite{ssnet}  &Supervised& $86.66 $ & $75.43 $ &$24.7$& $85.48 $ & $74.23 $ & $85.73 $ & $74.32 $ & $6.77 $ & $7.67 $ \\
  Wang's\cite{wangs}  &Supervised& $86.81 $ & $76.58 $ &$22.9$& $86.07 $ & $74.86 $ & $\underline{86.39}$ & $75.22 $ & $6.77 $ & $\underline{7.70} $ \\
  Pins\cite{Pins}  &Self-sup& $81.88 $ & $63.09 $ &$29.5$& $80.82 $ & $65.92 $ & $80.90 $ & $65.94 $ & $3.34 $ & $3.49 $ \\
  SL\cite{selflifting} &Self-sup&$\underline{86.70} $ & $\underline{76.96} $ &$\underline{22.6}$& $85.86 $ & $75.07 $ & $85.92 $ & $74.73 $ & $\underline{6.94} $ & $7.65 $ \\
  CMPC\cite{cmpc} &Self-sup& $86.54 $ & $76.01 $ &$23.3$& $\underline{85.91} $ & $\underline{75.24} $ & $85.50 $ & $\underline{75.82} $ & $6.41 $ & $6.75 $ \\
\midrule
  Ours &Self-sup& $\mathbf{89.14} $ & $\mathbf{79.98} $ &$\mathbf{19.3}$& $\mathbf{88.09} $ & $\mathbf{78.01} $ & $\mathbf{88.83} $ & $\mathbf{78.13} $ & $\mathbf{8.37} $ & $\mathbf{8.91} $ \\
\bottomrule
  \end{tabular}
\end{table*}

\subsection{Implementation details}
\textbf{Network architecture.}
We use the ECAPA-TDNN\cite{ecapatdnn} pre-trained on VoxCeleb2\cite{voxceleb2} as the voice encoder, which takes an 80-dimensional log mel-spectrogram from the voice clips as input and outputs a 192-dimensional embedding. Meanwhile, we employ the Inception-V1\cite{inceptionv1} pre-trained on VGGFace2\cite{cao2018vggface2} as the face encoder, processing input images reshaped into 3x112x112 and outputting a 512-dimensional embedding. To refine our model, the original fully-connected layer of both encoders is removed, and after obtaining the embeddings from the encoder outputs, we project them into a 256-dimensional space. Lastly, our multimodal encoder incorporates a 4-layer transformer\cite{transformer} block with 256 hidden size and 4 attention heads, ensuring a robust architecture.

\textbf{Training strategy.}
The training process is divided into two stages: clustering and model training. In the clustering stage, the initial number of clusters $C$ is 16650, the number of videos used for training. We use the verification metric as the evaluation criterion. If the verification performance on the validation set does not improve in the last 3 training epochs, we halve the number of clusters $C$. During the model training stage, we train FAA with the AdamW\cite{adamw} optimizer, using an initial learning rate of $10^{-4}$ , a weight decay of 0.01, and a batch size of 256. The maximum number of epochs is set to 50, and the best model on the validation set is preserved for evaluation. The other hyper parameters are set as follows: $M=256, \varepsilon=0.1,\alpha=2, \beta=40, \lambda=1, \delta=0.9$.

\textbf{Evaluation protocol.}
In order to comprehensively assess the performance of the model, we conduct the following protocols:
\begin{itemize}
    \item [1)] \textbf{Matching.} 
    The objective is to determine the candidate that corresponds with the probe. We are provided with a sample from one modality as the probe and two candidates from the other modality (one of which shares the same identity as the probe) as the gallery. We compare the binary classification probabilities produced by the multimodal encoder to find the result. The performance is evaluated using accuracy (ACC).

    \item [2)] \textbf{Verification.}
    The task involves determining if a given voice clip and face image belong to the same individual. We decide the result according to the binary classification result. The performance evaluation is conducted using the Area Under the ROC curve (AUC) or Error Rate (EER).
    
    \item [3)] \textbf{Retrieval.}
    If a probe from one modality is provided with a large gallery containing candidates from another modality, one or more candidates may correspond to the probe. The objective is to arrange the gallery so that the matched candidates are at the top of the ordering. Performance is evaluated using Mean Average Precision (mAP).
    
    Due to limited computational resources, we are unable to compute a binary classification score for an image against all other voices (vice versa) using a multimodal encoder. However, we can still use cosine similarity or L2 distance to find some candidates and then apply our multimodal encoder.

\end{itemize}


\begin{table*}[h]
  \centering
  \caption{Ablation study(\%) on  1:2 Matching and Retrieval.
}
  \label{tab:example}
  \begin{tabular}{c|c|c|cc|cc}
\toprule
  \multirow{2}{*}{Contrastive learning}&\multirow{2}{*}{Multimodal encoder} & \multirow{2}{*}{Effective Pair Selection}& \multicolumn{2}{c}{ 1:2 Matching(ACC) } & \multicolumn{2}{c}{ Retrieval(mAP) } \\
  &&& V2F(U)  & F2V(U)  & V2F & F2V \\
\midrule
  Contrastive Loss &\XSolid & \XSolid&$66.1 $  & $62.9 $ & $1.33 $ & $0.84 $ \\
  Contrastive Loss  &\Checkmark& \XSolid & $86.8 $  & $75.3 $ & $1.40 $ & $0.83 $ \\
  MultiSimilarity Loss  &\XSolid& \XSolid & ${85.81} $  & $85.97 $ & $8.12 $ & $8.65 $ \\
  MultiSimilarity Loss  & \XSolid &\Checkmark& ${86.36} $  & $86.54 $ & $8.29 $ & $8.78 $ \\
  MultiSimilarity Loss  &\Checkmark& \XSolid & $\mathbf{88.10} $  & $88.75 $ & $8.28 $ & $8.64 $ \\
  MultiSimilarity Loss  &\Checkmark&\Checkmark  & $88.09 $  & $\mathbf{88.83} $ & $\mathbf{8.37} $ & $\mathbf{8.91} $ \\
\bottomrule
  \end{tabular}
\end{table*}

\section{Results and comparison}

\subsection{comparison with the state-of-the-art frameworks}
\subsubsection{Quantitative results}
We now compare our FAA framework with previous works, including both supervised and unsupervised methods, and our model outperforms them all. To make a fair comparison, we use the same encoders for all our experiments. The results for Matching, Verification, and Retrieval are presented in Table 2. As we can see, our FAA achieves significant improvements, reaching the best performance in each task. In the Verification and Matching tasks, we observe an increase of around 3\%, while in the Retrieval task, we achieve an improvement of approximately 2\%.
\subsubsection{Ablation studies}
Table 3 presents the results of the ablation study focusing on the choice of contrastive learning loss, the use of a multimodal encoder, and the application of progressive clustering. For the Contrastive Learning loss function, we experiment with Contrastive Loss and Multi-similarity Loss. Regarding the use of the multimodal encoder in Matching and Verification tasks, when not using the multimodal encoder, we take the cosine similarity between the face embedding and voice embedding obtained from unimodal encoders as the evaluation criterion. When using the multimodal encoder, we use the pooled multimodal encoder's output embedding as the joint representation of the face-voice pair, followed by a fully connected layer and a softmax layer, resulting in a binary classification probability. As for progressive clustering, when not applying progressive clustering, we set the number of clusters to 1000. When using progressive clustering, we set the initial number of clusters to 16650, the number of videos used for training.

\begin{itemize}
    \item [1)] \textbf{Contrastive Learning Loss.}
    We first conduct ablation experiments on different loss functions. When using Contrastive Loss, the model can only measure the relationship between two samples, and we find that the model cannot converge, resulting in poor modality alignment. This is intuitively reflected in the lower Retrieval metric and indirectly leads to a significant gap between the V2F and F2V metrics in the 1:2 Matching task. However, when using MultiSimilarity Loss, the model can measure the relationships among multiple samples, which substantially improves the model, with the Retrieval metric increasing multiple times. Simultaneously, Both 1:2 Matching metrics (V2F and F2V) are all around 88\%, indicating better modality alignment as voice and face can be mapped to a similar vector space.

    \item [2)] \textbf{Multimodal Encoder.}
    We also conduct ablation experiments on the use of the multimodal encoder. We employ a 4-layer transformer block with 256 hidden size and 4 attention heads for the Verification and Matching tasks. As shown in Table 3, using the multimodal encoder can significantly improve the performance in the Matching task, even when the modality alignment is not optimal. This demonstrates that the multimodal encoder is more effective at learning cross-modal relations than cosine similarity or L2 distance. 

    Once we obtain the face embedding and voice embedding, if we only use cosine similarity for retrieval and matching tasks, we are merely treating the embeddings as high-dimensional vectors. However, we can learn the implicit information in the embeddings from a deeper and more diverse perspective by utilizing neural networks. This is more conducive to understanding the relationships between modalities.

    \item [3)] \textbf{Effective Pair Selection.}
    We perform progressive clustering and hard negative mining to find more diverse training pairs, hoping to enhance the effectiveness of Contrastive Learning. The results show that effective pair selection indeed brings a certain improvement in the Retrieval task, showing that the modality alignment performance is increased during the contrastive learning stage. However, in the Matching task, it is possible that due to the powerful cross-modal relation learning capability of the multimodal encoder, the improvements in modality alignment do not necessarily lead to better Matching performance.

\end{itemize}

\begin{figure}
  \centering
   \includegraphics[width=1\linewidth]{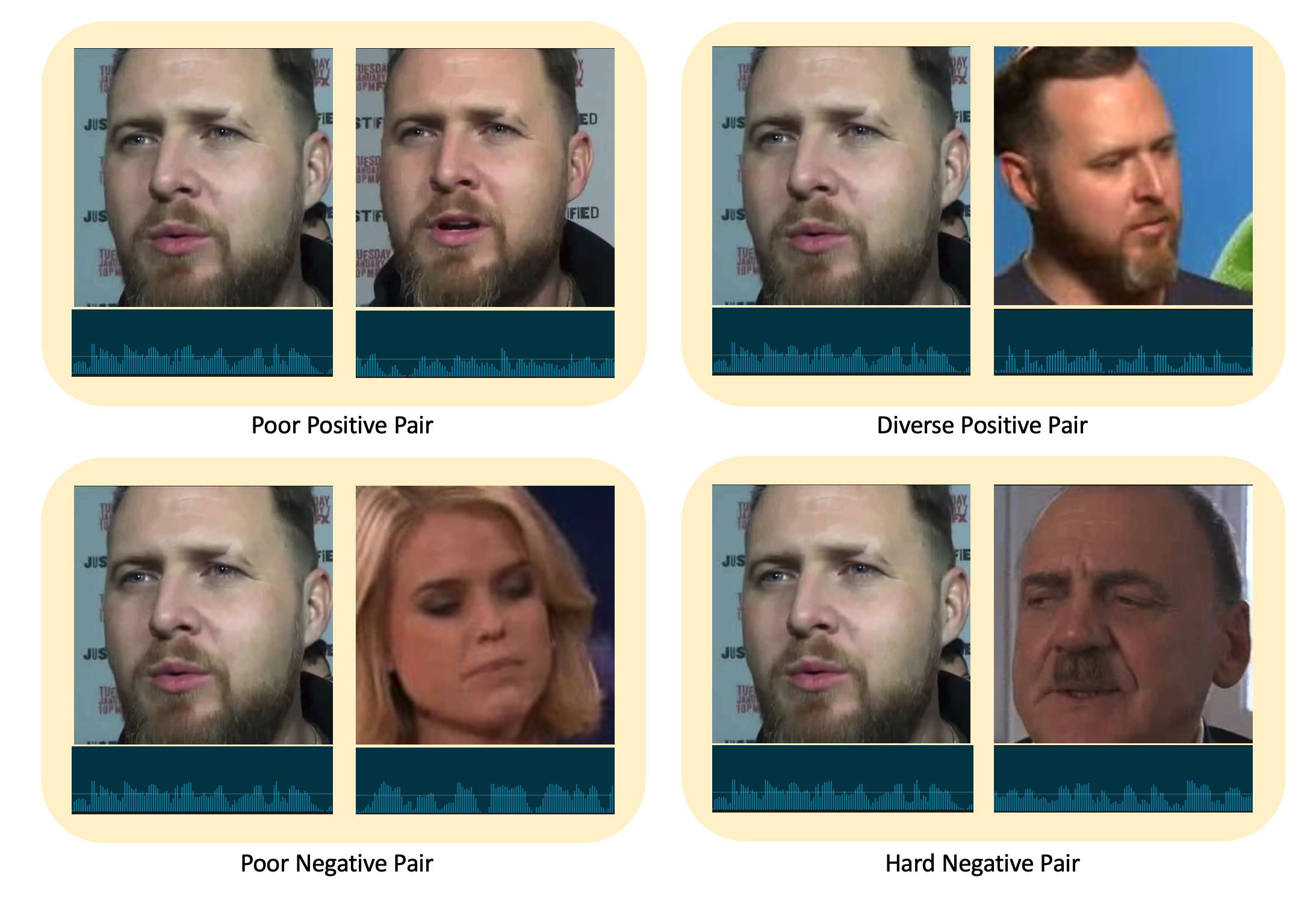}
   \caption{Examples of training pairs: poor positive pairs may be sampled from the same video segment, where they are very similar in aspects such as environment, attire, and facial expressions. In contrast, diverse positive pairs are sampled from the same individual but feature a variety of background elements. In the case of poor negative pairs, the difference between two distinct individuals can be very pronounced, making it easy for the model to discern that they are not the same person. However, hard negative pairs may involve two individuals who share certain similarities, such as gender and facial hair.}
   \label{fig:onecol}
\end{figure}

\section{Conclusion}
In this paper, we propose a novel framework for voice-face association learning. We introduce the use of a multimodal encoder to learn cross-modal relations more effectively. Additionally, we propose an effective pair selection method, which makes both positive and negative pairs more diverse and challenging, thereby enhancing the framework's capability to handle hard samples. Our experiment from tasks such as matching and retrieval proves the proposed framework's effectiveness. Compared to existing methods, the FAA outputs better results on multiple tasks. 
{\small
\bibliographystyle{ieee_fullname}
\bibliography{citation}
}

\end{document}